\documentclass[conference]{IEEEtran}
\IEEEoverridecommandlockouts
\usepackage{cite}
\usepackage{amsmath,amssymb,amsfonts}
\usepackage{algorithmic}
\usepackage{graphicx}
\usepackage{textcomp}
\usepackage{xcolor}
\usepackage{subfig}
\def\BibTeX{{\rm B\kern-.05em{\sc i\kern-.025em b}\kern-.08em
    T\kern-.1667em\lower.7ex\hbox{E}\kern-.125emX}}
\begin{document}

\title{Image to Language Understanding: Captioning approach}

\author{\IEEEauthorblockN{Madhavan Seshadri}
\IEEEauthorblockA{\textit{School of Engineering and Applied Science} \\
\textit{Columbia University}
}
\and
\IEEEauthorblockN{Malavika Srikanth}
\IEEEauthorblockA{\textit{School of Engineering and Applied Science} \\
\textit{Columbia University}
}\IEEEauthorblockN{Mikhail Belov}
\IEEEauthorblockA{\textit{School of Engineering and Applied Science} \\
\textit{Columbia University}
}
}

\maketitle

\begin{abstract}
Extracting context from visual representations is of utmost importance in the advancement of Computer Science. Representation of such a format in Natural Language has huge variety of applications such as helping the visually impaired etc. Such an approach is a combination of Computer Vision and Natural Language techniques which is a hard problem to solve. This project aims to compare different approaches for solving the image captioning problem. In specific, the focus was on comparing two different types of models: Encoder-Decoder approach and a Multi-model approach. In the encoder-decoder approach, inject and merge architectures were compared against a multi-modal image captioning approach based primarily on object detection. These approaches have been compared on the basis on state of the art sentence comparison metrics such as BLEU, GLEU, Meteor and Rouge on a subset of the Google Conceptual captions dataset which contains 100k images. On the basis on this comparison, we observed that the best model was Inception injected encoder model. This best approach has been deployed as a web based system. On uploading an image, such a system will output the best caption associated with the image.
\end{abstract}


\section{Introduction}
Extraction of visual information from images and videos have gained traction in the recent years both in the field of Computer Vision and Natural Language Processing. Image captioning is the extraction of the best description of the image in a natural language form. This has wide variety of applications such as helping the visually impaired people understand the contents of the image, automatic defect detection in supply chain pipelines, searching and indexing large quantities of images on the internet. This is regarded as the grand challenge in the field of computer science. Generating the best caption for an image can go above an beyond a simple image classification, object detection, context detection tasks.

This problem can be tackled using two different approaches, bottom-up where a set of words are generated from the image and are used to form a sentence and top-down in which a gist is automatically generated from the image into words. Bottom-up approaches are studied in detail in \cite{b1}, \cite{b2}, \cite{b3} and the top-down approaches are studied in \cite{b4}, \cite{b5} and \cite{b6}.

Deep learning approaches have been gaining traction in both Computer Vision and NLP where Convolutional Neural Networks (and variations) are used to better predict objects/features for the classification tasks \cite{b10}, \cite{b11}. LSTMs (and variations) have also become popular for sentence prediction tasks \cite{b7}, \cite{b8}, \cite{b9}. Consequentially, deep learning for image captioning is also being extensively studied \cite{b12} and \cite{b13}.

This paper aims at developing a framework for the comparison, analysis and scalable development of Image Captioning Approaches on any dataset. The structure of this paper is as follows: After the introduction section, a comparison of the state of the art approaches of the problem is covered in Literature Review; Section 3 covers the brief introduction of datasets used for this problem. Section 4 covers Methods and Architecture of the system to be able to tackle the problem at scale. Section 5 covers comparison metrics for evaluating various models followed by Results and Discussion in Section 6. Finally we conclude with Conclusion and Future Work in Section 7. 

\section{Literature Review}
Traditional approaches in Image Captioning involved generating a set of feature detectors such as Objects, context, words and phrases which was then stitched together using a language model to form sentences in a bottom-up approach. \cite{b1} tries to detect visual dependencies between features of importance in the images. \cite{b12} uses a combination of CNNs and deep bidirectional LSTMs for generating captions from images. As with any bidirectional LSTM, it is capable of generating the best sentence while keeping a balance between the past state (history) and the future state. The authors had added several data augmentation techniques such as image cropping, multi-scale and vertical mirror to prevent model over-fitting on MS COCO dataset. Performance analysis indicated that bidirectional LSTMs significantly outperformed other models for this task. 

A variational auto-encoder approach for this problem was first proposed in \cite{b15}. In this the authors used a combination of CNN as an encoder and Deep Generative Deconvolutional Network as a decoder for this task. The DGDN model was expected to learn the prediction for words sequences given the encoded image. The authors claim that the model can learn to predict sentences even when labels/captions are not available for the images thus making this a partially unsupervised approach.

In \cite{b16}, the authors vary from the traditional encoder-decoder approach for caption generation to experiment with a Reinforcement learning based "Policy network" and "Value Network" to generate captions for images. The policy network is responsible for suggesting the next word state given the current state and the value network is supposed to provide the best global look ahead policy. A visual semantic reward policy is used for this Deep Learning approach.

\section{Dataset Overview}
Two datasets were utilized while training our deep learning models: Flickr8k and Google's Conceptual Captions \cite{b21}. Flickr8k contains 8000 images, each with 5 reference captions. It was used for the initial development of our models to obtain the preliminary results. Google's Conceptual Captions dataset contains 3.3 million images, each with 1 reference caption, as mentioned in  \cite{b21}. It was decided to use 100,000 subset of Google's dataset due to the time constraints of our project to speed up the training process of the deep learning models, and to decrease the download time of the Google's Conceptual Captions dataset.\par
Dataset pre-processing involved resizing and normalizing the images from both Flickr8k and Google's Conceptual Captions datasets, and creating the image encoding from the resized images using state-of-the-art Convolutional Neural Networks. Also, the captions were pre-processed by creating the word-to-id dictionaries of all the words that appear in both datasets to encode the image descriptions. \par
We performed the preliminary analysis of both datasets by applying the dimensionality reduction techniques such as PCA and t-SNE to provide some insight into the datasets' content. In addition, K-Means clustering was performed on both datasets with k = 20 to group the images into their respective clusters as a part of the data visualization process. \par

\begin{figure}[h!]
    \includegraphics[width=\linewidth]{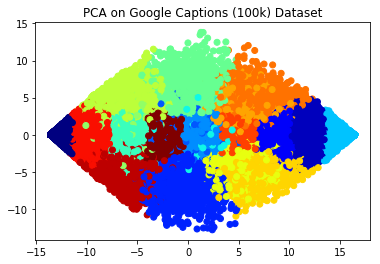}
    \caption{PCA on Google's Conceptual Captions Dataset}
 \end{figure}
 
\begin{figure}[h!]
    \includegraphics[width=\linewidth]{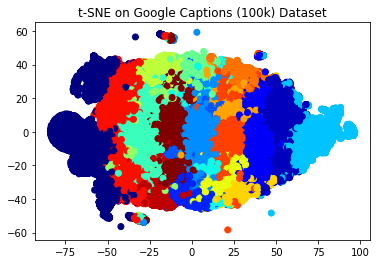}
    \caption{t-SNE on Google's Conceptual Captions Dataset}
 \end{figure}
 
As it is illustrated in PCA and t-SNE plots, there is lots of overlap between the images in both Flickr8k and Google's Conceptual Captions datasets. Moreover, no distinctive image clusters can be determined based on the obtained plots, which implies that deep learning approaches are optimal, given the high dimensionality and variance of the images in Flickr8k and Google's Conceptual Captions datasets.

\section{Model Architecture}
There are 2 types of approaches commonly used for image captioning - \\
1. Encoder-Decoder Architecture-Based approaches \\
2. Compositional Architecture based approaches \\

\subsection{Encoder-Decoder Architecture-Based approaches}
This image captioning method works in an end to end manner. In this method, image features are extracted from the hidden activation of the encoder model. These image features are then fed into the decoder which generates a textual description output.
Encoder-Decoder based approaches for image captioning can be implemented in 2 ways which differ primarily in where the image is places in the image caption generator. The 2 ways are the inject and merge based architectures. \\

\subsubsection{Inject Architecture} 
In the inject model (Fig. 3), the encoder first encodes the image into a fixed length vector representation. The model combines the encoded form of the image with each word from the text description generated so-far. The decoder acts as a text generation model that uses a sequence of both image and word information as input in order to generate the next word in the sequence. Thus, this is a form of image-conditioned language generation where the decoder generates sequences based on both linguistic and perceptual information.

 \begin{figure}[h!]
    \includegraphics[width=\linewidth]{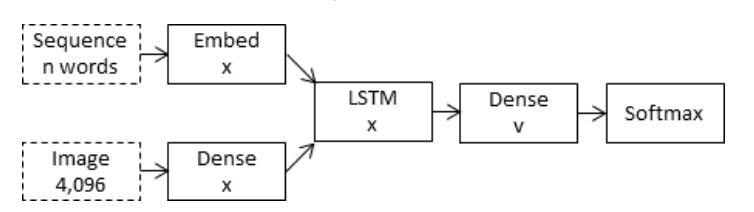}
    \caption{Inject Architecture}
 \end{figure}

We implemented the Inject architecture based model with a CNN as the Encoder and an LSTM as the decoder. We implemented the model with the following variations and compared the results - \\
1. Inception V3 model as the encoder and an LSTM decoder.
2. Resnet 50 model as the encoder and an LSTM decoder. \\

\subsubsection{Merge Architecture} 
The merge model (Fig. 4) combines both the encoded form of the image input with the encoded form of the textual description generated so far. The combination of these two encoded inputs is then used by a simple decoder model to generate the next word in the sequence. This separates the modeling the image input, the text input and the combining and interpretation of the encoded inputs. \\
Thus, the role of the LSTM in the Inject and Merge architectures is very different. In the Merge model, the LSTM handles only purely linguistic information and its purpose is to generate an encoding of the sequence so far. On the other hand, in the Inject model, the LSTM acts as a text generation module and generates text using both linguistic and perceptual information.

 \begin{figure}[h!]
    \includegraphics[width=\linewidth]{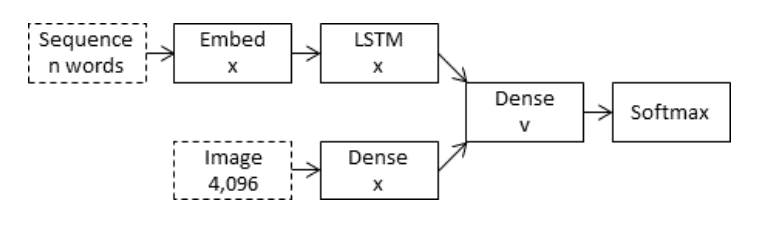}
    \caption{Merge Architecture}
 \end{figure}

We implemented the Merge architecture based model with a CNN as the image Encoder and an LSTM as the sentence encoder. We implemented the model with the following variations and compared the results - \\
1. Inception V3 model as the image encoder and an LSTM as the sentence encoder. \\
2. Resnet 50 model as the image encoder and an LSTM as the sentence encoder. \\

\begin{figure*}
  \includegraphics[width=\textwidth]{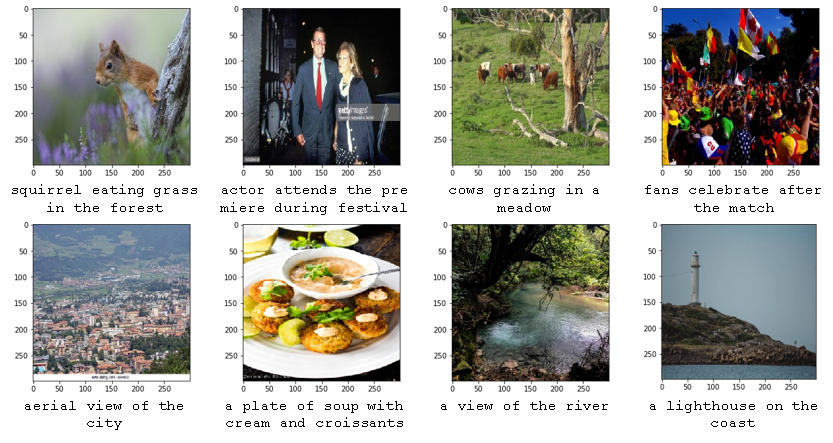}
  \caption{Sample Captions}
\end{figure*}

The Google Conceptual Captions dataset was cleaned and all the above models were run on the dataset with a token type minimum count of 4 (removing tokens with frequency less than 4). The models were also run on the dataset without any minimum count on the token type to assess the effectiveness of thresholding on the tokens.

\subsection{Multi-Modal Object Detection Architecture}
Multi-Modal object detection approach, that is partially based on \cite{b22}, involves extracting the entity information from a given image, and encoding that information using a specific language encoder. In our multi-modal approach, Inception (v.3) CNN was used to extract the top 5 most likely objects in the image, and, then, these objects were encoded into 512-dimensional vector using Google's Universal Sentence Encoder. \par
Our implementation of Multi-Modal Object Detection model follows the same encoder-decoder architecture described in the previous sections of the report. The key difference is that the decoder's output is now conditioned on the encoded objects detected in a given images, and not on the encoded image directly. \par
This approach explores the possibility that extracting the entity information from the images might result in a more accurate representation of the content in the images than what the standard image encoding allows us to capture. Moreover, the object encoding has 4 times less the dimensionality than the image encoding, which decreases the number of trainable parameters in the deep learning model by a factor of 0.9.

\subsection{Convolutional Neural Networks}
In our models, a Convolutional Neural Network was used to extract features from the image. In practice, it is rare to train an entire Convolutional Network from scratch, because it requires large datasets and high compute power. Instead, it is common to pre-train a CNN on a very large dataset, and then use this CNN for the task of interest. For our task, we worked with 2 different pre-trained CNN architectures - Inception-v3 and ResNet50.
\subsubsection{Inception-v3}
Inception-v3 (Fig. 6) is a convolutional neural network that has been trained on a large number of images in the ImageNet database. The network is 48 layers deep and can classify images into 1000 object categories. Inception-v3 has an image input size of 299-by-299.

 \begin{figure}[h!]
    \includegraphics[width=\linewidth]{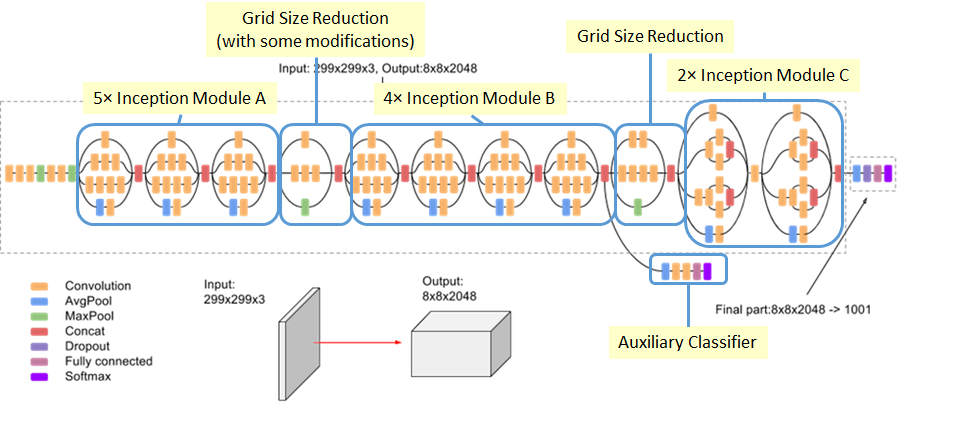}
    \caption{Inception-v3 Architecture}
 \end{figure}

\subsubsection{ResNet-50}
ResNet-50 (Fig. 7) is a CNN that is trained on a large number of images from the ImageNet database. The network is 50 layers deep and can classify images into 1000 object categories. ResNet-50 has an image input size of 224-by-224.

 \begin{figure}[h!]
    \includegraphics[width=\linewidth]{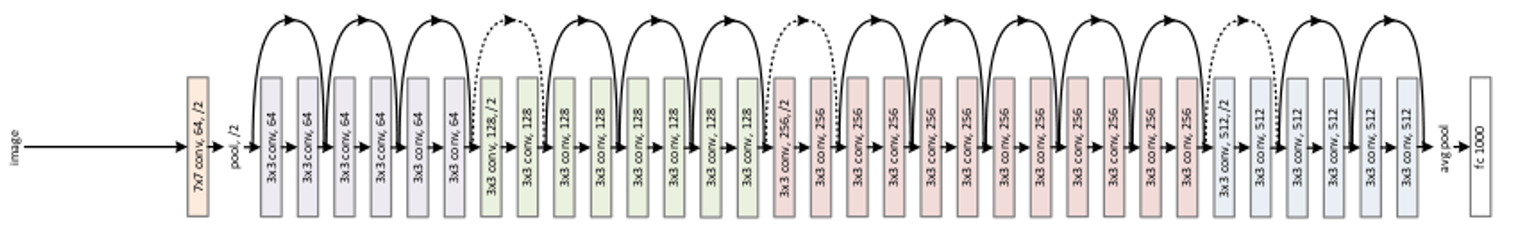}
    \caption{Resnet-50 Architecture}
 \end{figure}

\subsection{Long Short Term Memory Networks}
Our image captioning approaches use an LSTM (Fig. 7) for language modelling. The LSTM is used to predict the next word in a sentence in the Inject Architecture and it is used to generate an encoding of the sentence generated so far in the Merge Architecture. \\
The LSTM is a special kind of RNN which is capable of learning long term dependencies. An LSTM unit typically consists of a cell, an input gate, a forget gate and an output gate. LSTMs are widely used in language modelling tasks.

 \begin{figure}[h!]
    \includegraphics[width=\linewidth]{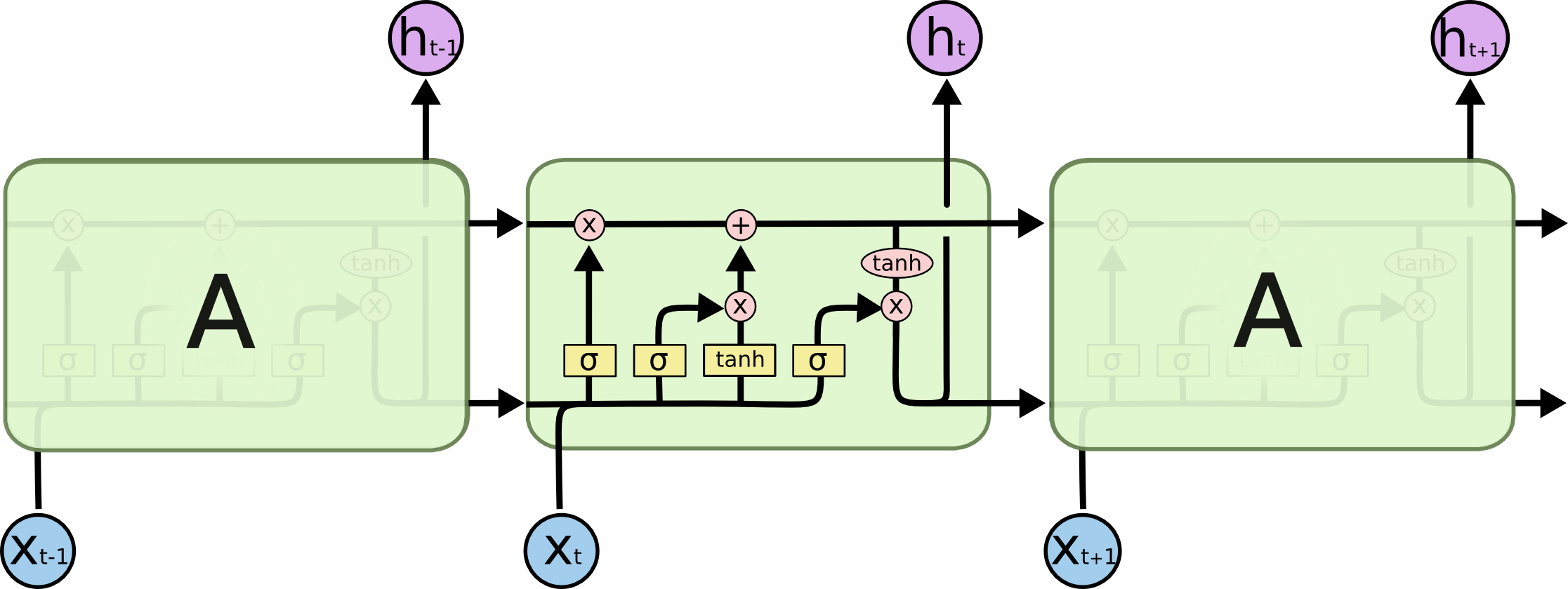}
    \caption{LSTM Architecture}
 \end{figure}
 
\section{Metrics}
In order to quantitatively assess our models' output image captions, we used 4 benchmark NLP metrics: BLEU, GLEU, METEOR, and ROUGE. We will discuss their definitions, and pros \& cons in the following subsections.

\subsection{BLEU}

BLEU is the most popular method for measuring the similarity between sentences, and widely used to assess the quality of the generated image captions. As defined in \cite{b14}, BLEU score is the geometric mean of n-gram precision scores multiplied by the brevity penalty for shorter sentences:
\begin{equation}
BLEU = min(1, \dfrac{candidate-length}{reference-length})(\prod_{i=1}^{4} Precision_{i})^{\dfrac{1}{4}}
\end{equation} \par
Even though BLEU is commonly used to assess the quality of image captions in the majority of the papers, it has some obvious limitations: it does not take n-gram recall into account, and word matching is not utilized when n-grams are being matched.

\subsection{GLEU}
GLEU metric is an improvement over BLEU score proposed by Google in their 2016 paper "Google’s Neural Machine Translation System: Bridging the Gap between Human and Machine Translation". This metric is used to evaluate the quality of the machine translation systems, but it can be used for evaluating the quality of image captions as well. \par
GLEU score is defined as the minimum of n-gram recall and precision, and it is calculated in the following manner as mentioned in \cite{b9}: "We record all sub-sequences of 1, 2, 3 or 4 tokens in output and target sequence (n-grams). We then compute a recall, which is the ratio of the number of matching n-grams to the number of total n-grams in the target (ground truth) sequence, and a precision, which is the ratio of the number of matching n-grams
to the number of total n-grams in the generated output sequence."

\subsection{METEOR}
METEOR, just as GLEU, is designed for the evaluation of the quality of machine translation systems. \cite{b14} defines it as the harmonic mean of precision and recall of unigram matches between sentences:
\begin{equation}
METEOR = F_{mean}(1-p)
\end{equation}
\par This metric addresses the disadvantages of BLEU by utilizing Word-Net to incorporate synonym matching, which partially resolves the problem of semantic similarity detection during the image caption quality evaluation.

\subsection{ROUGE}
ROUGE is initially proposed for evaluation of summarization systems, and this evaluation is done via comparing overlapping n-grams, word sequences and word pairs as noted by \cite{b14}. ROUGE measures the longest common sub-sequences between a pair of sentences, and it tends to favor longer and more complex sentences, despite their possible subject matter and semantic disparities.

\section{System}
A full fledged application was developed which takes an image as an input and outputs a caption associated with it. This is built in a modular manner with reusable components as depicted in Fig. 9. Since our deep learning model consists of an encoder-decoder pipeline. A CNN encoder model is used with pre-trained weights on imagenet. For the decoder, a trained Bidirectional LSTM is used. These weights are stored in a persistent disk and are loaded on system startup. A Django based application receives and processes the request which it then relays the message to an encoder model CNN encoding fed into Language model for sentence prediction. In a typical run, this process takes about 40 seconds.
 \begin{figure}[h!]
    \includegraphics[width=\linewidth]{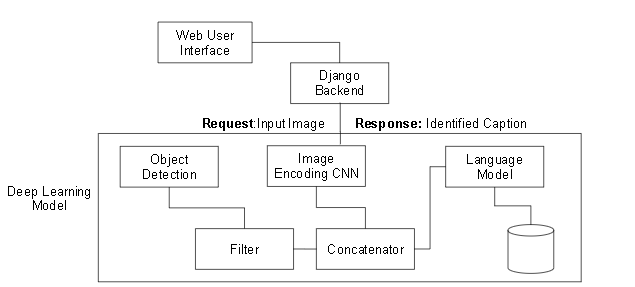}
    \caption{System Design}
 \end{figure}

While training deep learning models, we realised that in the process of choosing the best model for the task, several models were needed to be built and compared against on a common metric. This was achieved using a clean pipelined design for the Training process which is depicted in Fig. 10. In a plug-and-play architecture, any component can be added/replaced to compare. As explained before, we experimented with various deep learning architecture variations (inject and merge) and have paved a way to experiment with several more approaches such as a Teacher-Student architecture, complete auto-encoder architecture etc. This can be implemented with the system design with minimal code changes. On every training run, performance metrics such as Training time, training loss, training accuracy, validation loss and validation accuracy is also collected. We enabled early stopping to prevent the model from over-fitting the dataset and minimize training time. 
 
\begin{figure}[h!]
    \includegraphics[width=\linewidth]{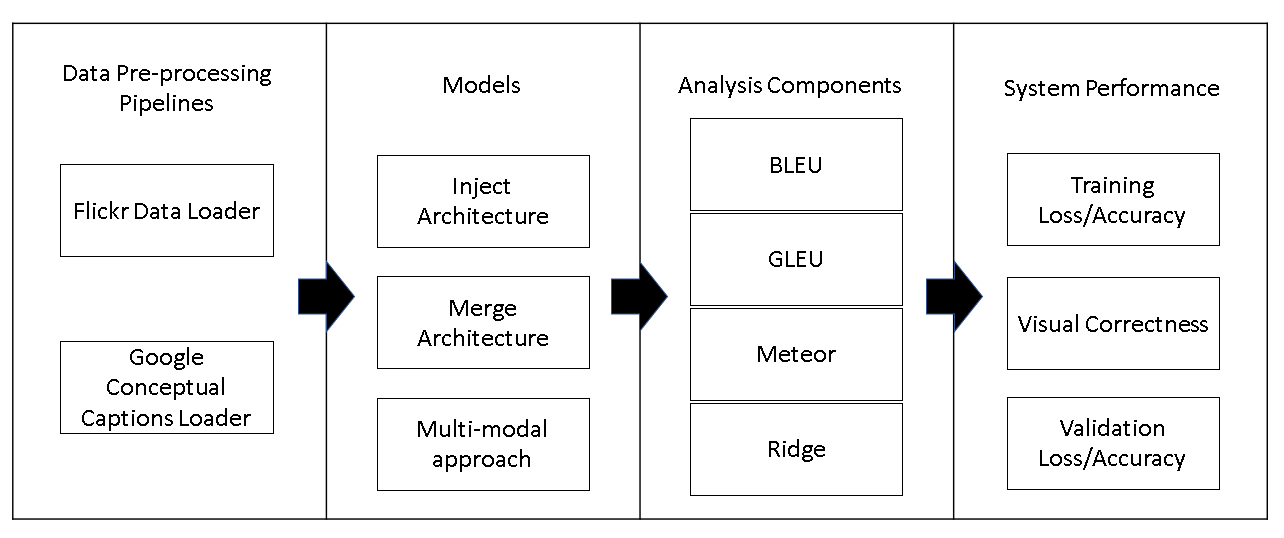}
    \caption{Deep Learning Training Pipeline}
\end{figure}

A snapshot of the UI is included in Fig. 11 and some generated captions are included in Fig 5. The tech stack used is Django, Python, Shell, Tensorflow, Keras, Linux, HTML, CSS, Javascript. The system can be further enhanced by employing Distributed training to speed up the computation time and increase the generalization by training on the complete dataset containing 3.3 million images.

\begin{figure}[h!]
    \includegraphics[width=\linewidth]{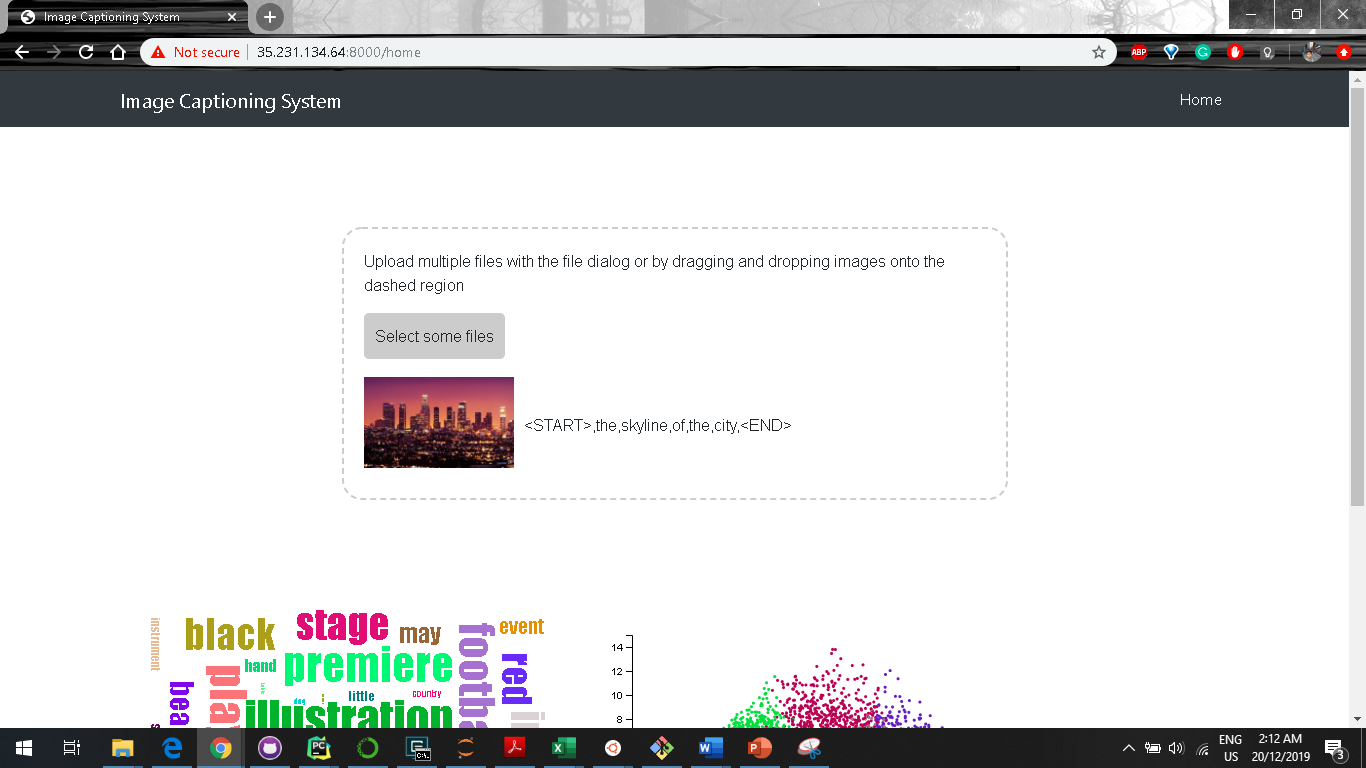}
    \caption{System Screenshot}
\end{figure}

\section{Results and Analysis}
Our results include the graphs of the training process of all the implemented deep learning models. Models' comparison is primarily based on the NLP metrics introduced in the "Metrics" section of the project report. 

\begin{figure*}[ht!]
   \subfloat{%
      \includegraphics[width=0.3\textwidth]{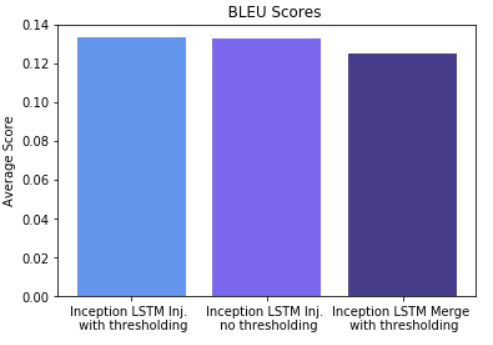}
      }
\hspace{\fill}
   \subfloat{%
      \includegraphics[width=0.3\textwidth]{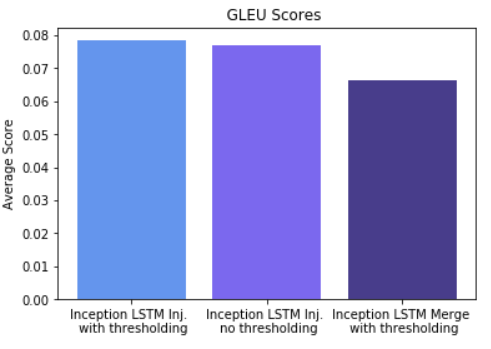}}
\hspace{\fill}
   \subfloat{%
      \includegraphics[width=0.3\textwidth]{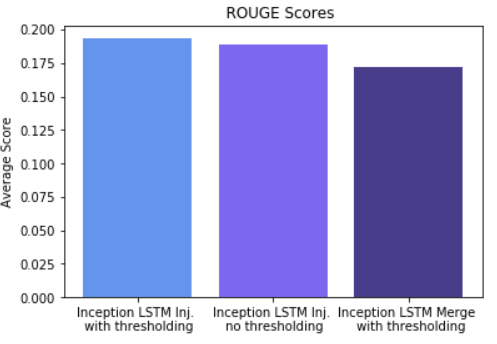}}\\
\caption{\label{workflow}Comparison between Inception LSTM models. (a) BLEU; (b) GLEU; (c) ROUGE }
\end{figure*}

\begin{figure*}[ht!]
   \subfloat{%
      \includegraphics[width=0.3\textwidth]{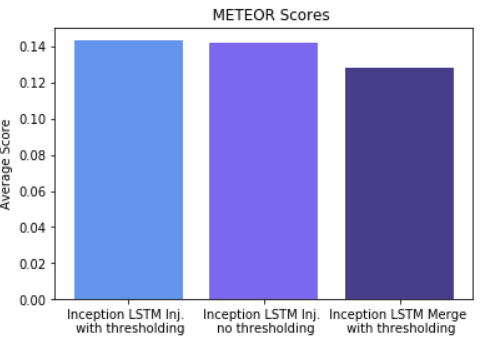}
      }
\hspace{\fill}
   \subfloat{%
      \includegraphics[width=0.3\textwidth]{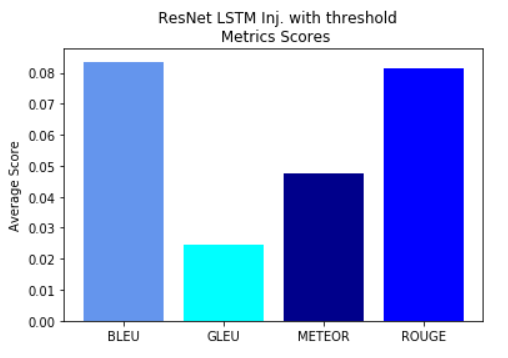}}
\hspace{\fill}
   \subfloat{%
      \includegraphics[width=0.3\textwidth]{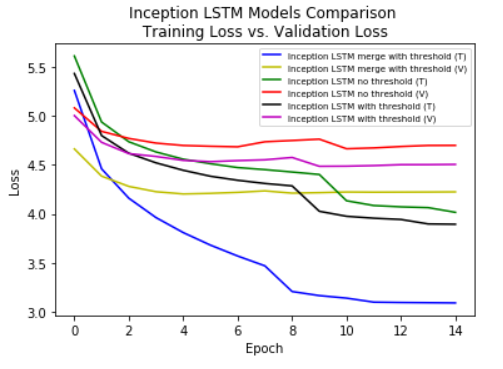}}\\
\caption{\label{workflow} (a) METEOR Comparison between Inception LSTM models; (b) ResNet LSTM Inject model without threshold; (c) Inception LSTM models: Training vs. Validation Loss }
\end{figure*}

\begin{figure*}[ht!]
   \subfloat{%
      \includegraphics[width=0.3\textwidth]{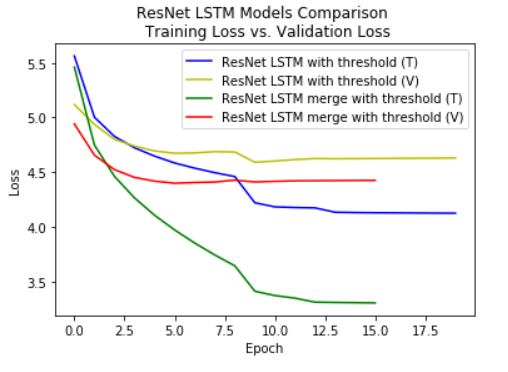}
      }
\hspace{\fill}
   \subfloat{%
      \includegraphics[width=0.3\textwidth]{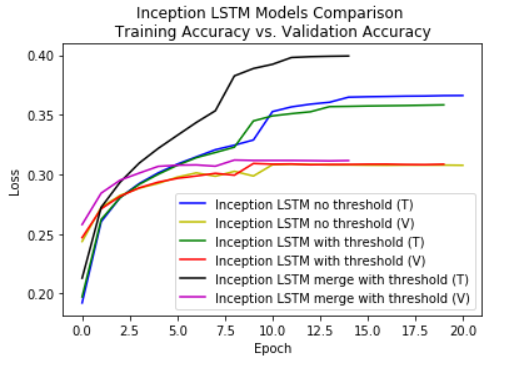}}
\hspace{\fill}
   \subfloat{%
      \includegraphics[width=0.3\textwidth]{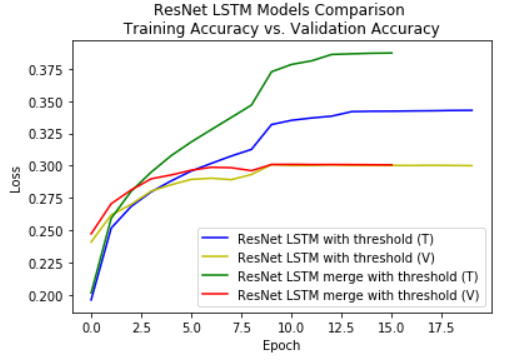}}\\
\caption{\label{workflow} (a) ResNet LSTM models: Training vs. Validation Loss; (b) Inception LSTM models: Training vs. Validation Accuracy; (c) ResNet LSTM models: Training vs. Validation Accuracy }
\end{figure*}

The graphs of the training process of each model in Fig. 14 provide us with some insight into how fast each model reached its convergence, and how the models can be further improved. The models were tested on a small subset of the images from the original Google Conceptual Captions training dataset. 
\par
All the models that we implemented tend to overfit relatively quickly (around the second epoch during training, Fig. 14), since the training accuracy decreases faster than the validation accuracy after the second epoch. Even though this undesirable behaviour is consistent across all the models, it is something to be expected when dealing with such a complex task, and training the model with only 100,000 images.\par
Meanwhile, there is no significant difference between the training process metrics of the models of the classic encoder-decoder architecture, we observed that Inception LSTM merge with threshold model had the lowest training loss and highest training and validation accuracy, while Inception LSTM Inject model had the best validation loss values. Therefore, we can conclude that, overall, Inception LSTM merge with threshold architecture is the optimal architecture in terms of the deep learning model metrics.\par
Also, NLP metrics in the form of the average BLEU, GLEU, ROUGE, and METEOR scores in Fig. 12 and 13 give us a tangible assessment of each model's performance. Since BLEU and METEOR scores are the most popular and widely used metrics, it is fair to put more weight onto these two metrics when assessing our models' performances.\par
Overall, we can conclude that Inception LSTM Inject with threshold architecture is the optimal deep learning model with BLEU, METEOR and ROUGE scores of 0.13, 0.14, and 0.18, respectively, in terms of the NLP metrics. Since this model had the highest validation accuracy, our expectations were that it would be the best model out of the ones implemented, and our prediction regarding this model's performance on the test set of our data was correct. \par
In addition, Multi-Modal Object Detection Architecture did not produce better results than the standard encoder-decoder architecture. This architecture had the highest training accuracy of 0.36 and validation accuracy of 0.25, and its training process had the same issue of overfitting after the second epoch as other encoder-decoder models.
\begin{figure}[h!]
    \includegraphics[width=\linewidth]{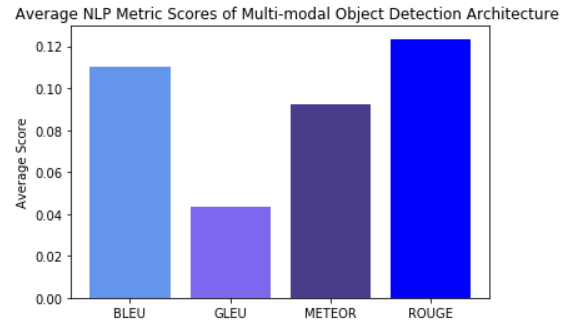}
    \caption{NLP Metrics of Multi-modal Object Detection Architecture}
\end{figure}
In terms of the NLP metrics (Fig. 15), Object Detection Architecture was not optimal as well. Such suboptimal results of this architecture can be attributed to the 4 times smaller encoding size (512-dimensional encoding size) that was used to condition the LSTM language model, which made it harder for the model to find the correlations between the content of images and the ground truth captions. \par
It is important to note that applying thresholding resulted in a slightly better performances across both training performance values and NLP metrics, which can be observed on the graphs in Fig. 12, 13, and 14. Thresholding reduces the vocabulary size, and, consequently, makes it easier to train the language model. The positive effect of thresholding is proved empirically by our final results, and it is something to incorporate in the improved versions of our implemented models. \par
Finally, we would like to point out that using ResNet as the image encoder in our implementation of the encoder-decoder architectures did not produce better results based on both training and NLP metrics when compared to Inception (v.3) CNN that we used in the majority of our models as the image encoder. \par

\section{Conclusion and Future Work}
This paper discussed in detail an architecture for building, training and validation a deep learning model for Image Captioning problem on a reasonably large dataset. Two architecture variations in the encoder-decoder approach (inject and merge) were compared on four main metrics BLEU, GLEU, Rouge and Meteor. Other system level performance metrics were also collected. A natural extension to this work would be to train on the complete Google Captions Dataset containing 3.3 million images using TF2.0 Distributed training approaches which can speed up the computation and training time. A complete auto-encoder approach can also be implemented.

\section*{Individual Contribution}
All authors contributed equally.

Video: https://youtu.be/GcrNIS9qr8Q\\
GitHub: https://github.com/MADHAVAN001/image-captioning-approaches


\begin{thebibliography}{00}
\bibitem{b1} D. Elliott and F. Keller. Image description using visual dependency representations. In EMNLP, pages 1292–1302, 2013.

\bibitem{b2} H. Fang, S. Gupta, F. Iandola, R. Srivastava, L. Deng, P. Dollar, J. Gao, X. He, M. Mitchell, J. Platt, et al. From captions to visual concepts and back. In CVPR, pages 1473–1482, 2015.

\bibitem{b3} P. Kuznetsova, V. Ordonez, A. C. Berg, T. L. Berg, and Y. Choi. Collective generation of natural image descriptions. In ACL, pages 359–368, 2012.

\bibitem{b4} X. Chen and C. L. Zitnick. Mind’s eye: A recurrent visual representation for image caption generation. In CVPR, pages 2422–2431, 2015.

\bibitem{b5}  J. Mao, W. Xu, Y. Yang, J. Wang, Z. Huang, and A. Yuille. Learning like a child: Fast novel visual concept learning from sentence descriptions of images. In ICCV, 2015.

\bibitem{b6} K. Xu, J. Ba, R. Kiros, A. Courville, R. Salakhutdinov, R. Zemel, and Y. Bengio. Show, attend and tell: Neural image caption generation with visual attention. In ICML, 2015.

\bibitem{b7} Ghosh, S., Vinyals, O., Strope, B., Roy, S., Dean, T., $\&$ Heck, L. (2016). Contextual lstm (clstm) models for large scale nlp tasks. arXiv preprint arXiv:1602.06291.

\bibitem{b8} Pichotta, K., $\&$ Mooney, R. J. (2016). Using sentence-level LSTM language models for script inference. arXiv preprint arXiv:1604.02993.

\bibitem{b9} Wu, Yonghui. “Google's Neural Machine Translation System: Bridging the Gap between Human and Machine Translation.” Https://Arxiv.org/, 8 Oct. 2016, https://arxiv.org/abs/1609.08144.

\bibitem{b10} Zhang, N., Donahue, J., Girshick, R., $\&$ Darrell, T. (2014, September). Part-based R-CNNs for fine-grained category detection. In European conference on computer vision (pp. 834-849). Springer, Cham.

\bibitem{b11} Szegedy, C., Ioffe, S., Vanhoucke, V., $\&$ Alemi, A. A. (2017, February). Inception-v4, inception-resnet and the impact of residual connections on learning. In Thirty-First AAAI Conference on Artificial Intelligence.

\bibitem{b12} Wang, C., Yang, H., Bartz, C., $\&$ Meinel, C. (2016, October). Image captioning with deep bidirectional LSTMs. In Proceedings of the 24th ACM international conference on Multimedia (pp. 988-997). ACM.

\bibitem{b13} Pu, Y., Gan, Z., Henao, R., Yuan, X., Li, C., Stevens, A., $\&$ Carin, L. (2016). Variational autoencoder for deep learning of images, labels and captions. In Advances in neural information processing systems (pp. 2352-2360).

\bibitem{b14} Kilickaya, Mert. “Re-Evaluating Automatic Metrics for Image Captioning.” Https://Www.aclweb.org/, Association for Computational Linguistics, Apr. 2017, https://www.aclweb.org/anthology/E17-1019/.

\bibitem{b15} Pu, Y., Gan, Z., Henao, R., Yuan, X., Li, C., Stevens, A., $\&$ Carin, L. (2016). Variational autoencoder for deep learning of images, labels and captions. In Advances in neural information processing systems (pp. 2352-2360).

\bibitem{b16} Ren, Z., Wang, X., Zhang, N., Lv, X., $\&$ Li, L. J. (2017). Deep reinforcement learning-based image captioning with embedding reward. In Proceedings of the IEEE Conference on Computer Vision and Pattern Recognition (pp. 290-298).

\bibitem{b17} Jason Brownlee (2017), Caption Generation with the Inject and Merge Encoder-Decoder Models. Retrieved from https://machinelearningmastery.com/caption-generation-inject-merge-architectures-encoder-decoder-model/

\bibitem{b18} Szegedy, C., Vanhoucke, V., Ioffe, S., Shlens, J., $\&$ Wojna, Z. (2016). Rethinking the inception architecture for computer vision. In Proceedings of the IEEE conference on computer vision and pattern recognition (pp. 2818-2826).

\bibitem{b19} He, K., Zhang, X., Ren, S., $\&$ Sun, J. (2016). Deep residual learning for image recognition. In Proceedings of the IEEE conference on computer vision and pattern recognition (pp. 770-778).

\bibitem{b20} Colah (2015), Understanding LSTM Networks. Retrieved from https://colah.github.io/posts/2015-08-Understanding-LSTMs/

\bibitem{b21} Sharma, Piyush. “Conceptual Captions: A Cleaned, Hypernymed, Image Alt-Text Dataset For Automatic Image Captioning.” Https://Www.aclweb.org/, Association for Computational Linguistics, 2018, https://www.aclweb.org/anthology/P18-1238/

\bibitem{b22} Fang, H. (2014, November 18). From Captions to Visual Concepts and Back. Retrieved from https://arxiv.org/abs/1411.4952v3.



\end{thebibliography}
\end{document}